\pdfoutput=1
\documentclass[11pt]{article}

\usepackage[]{naacl2021}

\usepackage{times}
\usepackage{latexsym}

\usepackage[T1]{fontenc}
\usepackage[utf8]{inputenc}

\usepackage{microtype}
\usepackage{xcolor}
\usepackage{amsmath}
\usepackage{amssymb}
\usepackage{wrapfig}
\usepackage{graphicx}
\usepackage{booktabs}

\newcommand{\ex}[1]{\textit{#1}}
\newcommand{\fm}[1]{\textsl{#1}}

\newcommand{\conc}[1]{\textsf{#1}}
\newcommand{\op}[1]{\textsc{#1}}
\newcommand{\att}[1]{\texttt{#1}}

\title{Jointly Learning Truth-Conditional Denotations and Groundings
  using Parallel Attention}

\author{Leon Bergen  \\
  UCSD \\
  {\small \texttt{lbergen@ucsd.edu} } \\\And
  Dzmitry Bahdanau  \\
  ServiceNow \\
  {\small \texttt{dimabgv@gmail.com}} \\ \And
  Timothy J. O'Donnell  \\
  McGill \\
  {\small \texttt{timothy.odonnell@mcgill.ca }} \\}

\begin{document}
\maketitle
\begin{abstract}
  We present a model that jointly learns the denotations of words
  together with their groundings using a truth-conditional
  semantics. Our model builds on the \fm{neurosymbolic} approach of
  \citet[][]{mao.j:2019}, learning to ground objects in the CLEVR
  dataset \citep[][]{johnson.j:2017} using a novel parallel attention
  mechanism.  The model achieves state of the art performance on visual
  question answering, learning to detect and ground objects with
  question performance as the only training signal. We also show that the
  model is able to learn flexible non-canonical groundings just by
  adjusting answers to questions in the training set. 
\end{abstract}

\section{Introduction}
The meaning of words and phrases is very flexible. An expression like
\ex{wooden table} can refer to an enormous variety of styles of table,
kinds of wood, etc. and can be used across a large variety of
different conversational contexts and situations.  To capture this
flexibility, many theories of semantics break down word meaning into
two parts. First, word \fm{denotations} capture situation-independent
aspects of meaning. The denotation of \ex{table}, for instance, might
encode features that are associated with tablehood---like having a
broad flat surface. In truth conditional semantics, denotations are
often modeled using boolean-valued predicates like
\textsc{table}($\cdot$) \citep[see,
e.g.,][]{heim.i:2000,mcconnell-ginet.s:2000}. Second, \fm{groundings}
associate words with information in some specific situation
\citep[][]{harnad.s:1990}.  A grounding for \ex{table}
could indicate, amongst other things, what region of space in a room
is occupied by the table in question. Groundings are often modeled by
the binding of situation-specific information to \fm{grounding
  variables} which are passed to denotations, for example, $x$ in
\textsc{table}($x$). 
%\footnote{Code for the paper is available at the following link: https://github.com/canonical-grounding/learning-groundings}

Denotation and grounding are tightly linked. For instance, while a
region of space, a visual texture, or a particular configuration of
shapes are appropriate kinds of information for evaluating tablehood,
different information is appropriate for evaluating the meaning of a
spatial relation such as \ex{on top}. Groundings are flexible and
influenced by language. For example, the denotations
of individual words can require grounding in collections of objects
(e.g., \ex{a forest}). Groundings can also be influenced by social
context.  The objects depicted in a painted mural are relevant during
a lecture on art history, but will be ignored when discussing how to rearrange
furniture in a room. 

Recent years have seen a great deal of work in language grounding in
domains such as images and video; however, most of this work does not
make use of the truth-conditional framework with its advantages in
modularity and compositionality (see
\textsection\ref{sec:background}). At the same time, little work from
within linguistic semantics has addressed the problem of learning the
complex relationships between denotations and groundings just
mentioned.

In this paper, we address this gap, studying the problem of
simultaneously learning the meaning of words and their groundings in
a truth-conditional setting. Building on the work of
\citet[][]{mao.j:2019}, we show how to use a novel \fm{parallel
  attention} mechanism to bind perceptual information to grounding
variables representing visual objects in the CLEVR dataset
\citep[][]{johnson.j:2017}.

Our system achieves state-of-the-art performance on the visual
question answering task (VQA).  Critically, unlike
\citet[][]{mao.j:2019} our model is trained end-to-end and makes no
use of an external object-recognition module. Most importantly, the
model is able to learn how to bind grounding variables to percepts by
back-propagating entirely through the compositional semantics, with no
other external training signal. We perform a number of ablation
experiments to better understand which components of our model are
critical.  Finally, we demonstrate that our model can capture the
top-down influence of semantics; we show that the model can acquire
non-canonical groundings just by adjusting its linguistic input.

\section{Background}
\label{sec:background}
Formal semantics formalizes word
meanings using \fm{predicates} of variables that encode situational
groundings \citep[][]{lakoff.g:1970,montague.r:1970}.  In Montague semantics, these predicates capture truth conditions on
word meaning
\citep[][]{heim.i:2000,mcconnell-ginet.s:2000,gamut.l:1991}. However,
the use of word-meaning predicates (or functions) is common across
many other varieties of semantics \citep[e.g.,
][]{pietroski.p:2005,jackendoff.r:1990}. The advantages of an
architecture which distinguishes denotations and groundings in this
way is that (i) it provides a
clean division of labor between situation-specific (grounding) and
situation-general (denotation) aspects of the problem and (ii) it
allows for compositionality, in which the meaning of
composed expressions can be built up from logical combinations of
functions of grounding variables. For example the meaning of
\ex{wooden table} can be composed as \textsc{wooden}($x$) $\wedge$
\textsc{table}($x$).

Linguistic semantics has not historically focused on the
problem of how grounding variables get bound to information from
outside the linguistic system.
% DIMA: researchers in robotics and computer vision did not seek to ground the meaning of expressions. They just wanted to build systems that allow asking questions about images or giving robots natural language commands. Language grounding for many such researchers is the means for achieving the goal, not the goal itself. 
By contrast, the last years have seen an explosion of work that
tackles this problem of\fm{grounded language learning}. This includes
linguistically inspired grounding models from computational
linguistics and AI
\citep{matuszek_joint_2012,krishnamurthy_jointly_2013,yu.h:2015,siskind.j:1992}
or robotics \citep[][]{kollar.t:2013,tellex.s:2014}, computer vision
models that answer questions about images,
\citep{antol_vqa:_2015,geman_visual_2015} or videos
\citep{tapaswi_movieqa_2016}, and reinforcement learning models of how
grounded meaning can be learned by trial and error in a simulated 3D
world \citep[][]{hermann_grounded_2017}.

Early work in grounded language learning made use of predicate-based
representations of meaning similar to those used in linguistics
\citep[e.g.,][]{siskind.j:1992}, and probabilistic variants of these
approaches continue to be explored in AI
\citep[e.g.,][]{matuszek_joint_2012,yu.h:2015,krishnamurthy_jointly_2013}
and robotics \citep[e.g.,][]{tellex.s:2014,kollar.t:2013}. However,
these approaches are typically non-differentiable and, for this
reason, most current work in the area makes use of jointly trained
neural models
\citep{perez_film:_2017,hudson_compositional_2018,anderson_bottom-up_2018}.
%\todo{Do we want to discuss NMNs as the special case} \tjo{I think that NMNs are an attempt to compromise between the two approaches in a way. the modules are meant to be like functions }. 
Although some of these models do feature dedicated word-specific neural functions
\citep[that is, \fm{modules}, see
e.g.,][]{andreas_neural_2016,johnson_inferring_2017}, these functions
are not predicates in the usual sense. Instead, they usually return distributed or
attention-based representations of word meaning which can limit their
ability to generalize systematically
\citep{bahdanau_closure_2019}.

One recent strand of work---the \fm{neurosymbolic} approach of
\citet{mao.j:2019}---attempts to combine the truth-conditional and
neural approaches to grounded language learning. By making use of a
differentiable variant of truth-conditional semantics---roughly, a
fuzzy-logic-based semantics---\citet{mao.j:2019} show how meanings can
be learned in a predicate-based but fully differentiable
system. \citet{mao.j:2019} are able to achieve near-perfect
performance on the CLEVR visual question answering dataset.

There is one important limitation of the work of
\citet{mao.j:2019}. While their model learns denotations of individual
words, it does not jointly learn groundings for those words. Instead,
the model makes use of an out-of-the-box, pretrained object recognizer
to find bounding boxes around candidate objects in images. These
bounding boxes are then used to derive the featural representations of
objects which are bound to grounding variables in the model. Because
these bounding boxes are provided by an external system, they cannot
be influenced by the linguistic system.

In this work, we extend the model of \citet{mao.j:2019} to jointly
learn denotations and groundings through a novel mechanism which we
call \fm{parallel attention}. The task is closely related to recent work on self-supervised learning of object representations \citep{burgess2019monet, engelcke2019genesis, greff2019multi, greff2016tagger, jiang2020scalor, lin2020space,locatello2020object}. That work aims to learn object representations without bounding box supervision, though the training signal differs from the current work, using reconstruction loss instead of linguistic signal. We describe our model in the next section.

\begin{figure*}
    \includegraphics[width=0.8\paperwidth]{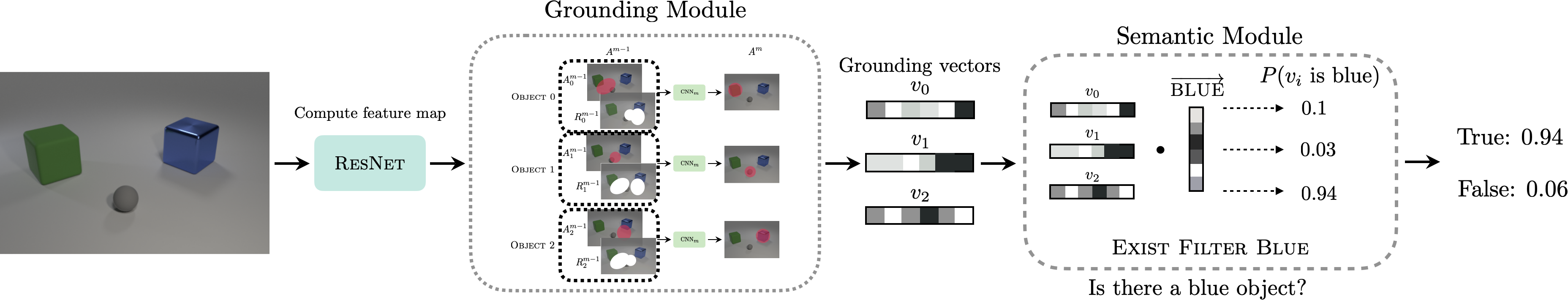}
    \caption{Model architecture. During training, the model's distribution over predicted answers is compared to the ground truth answer; this provides the only training signal. The model assumes access to ground truth semantic parses.}
    \label{fig:architecture-overview}
\end{figure*}

\begin{figure}
    \includegraphics[width=0.35\paperwidth]{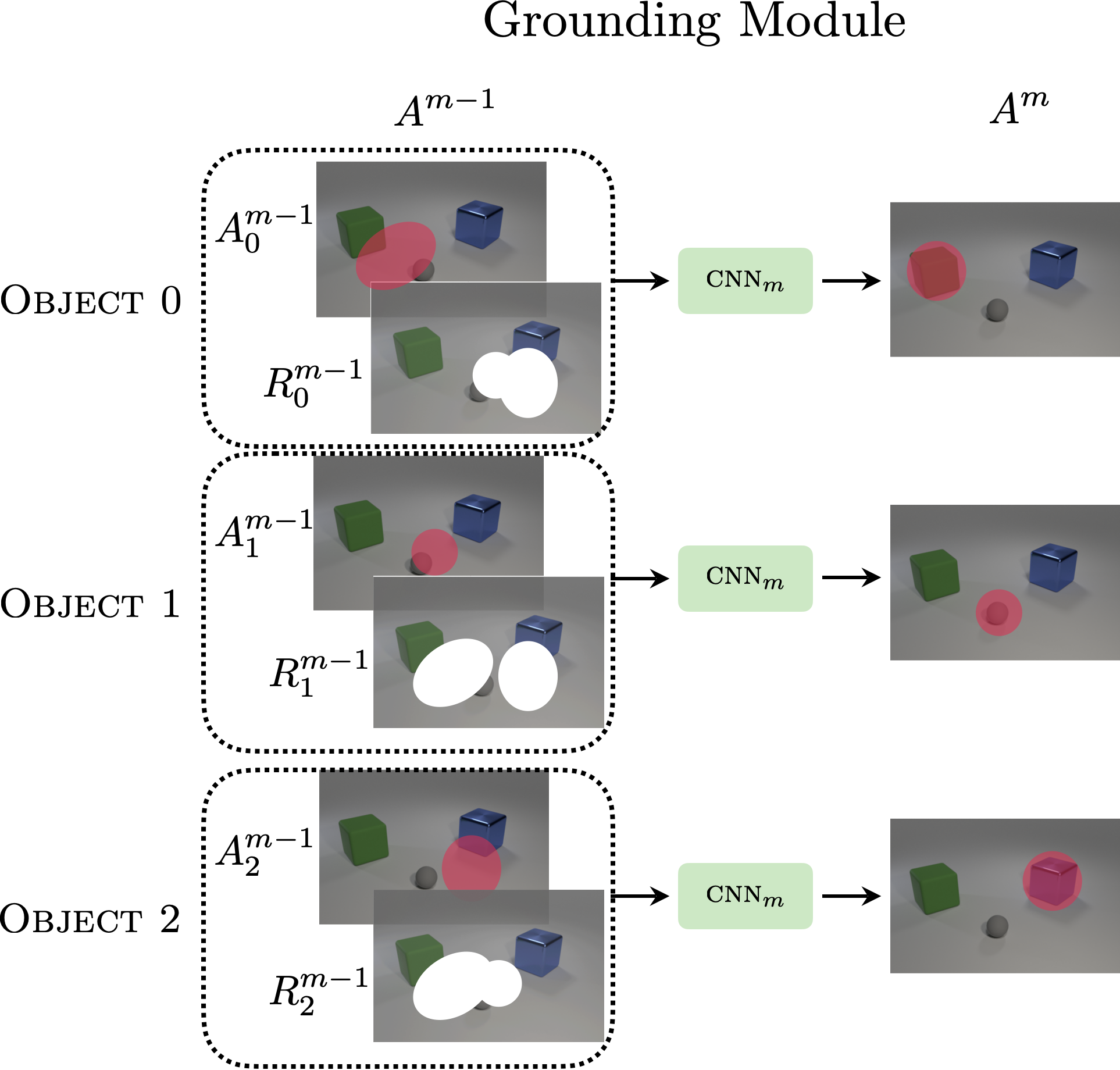}
    \caption{Grounding module from Figure \ref{fig:architecture-overview}, showing one step in the refinement of attention maps. Red indicates location of attention, white indicates location of scope. The number of objects in a scene is learned with the objecthood parameter $h_i$ defined in Equation \ref{eq:objecthood}.}
    \label{fig:attention-illustration}
\end{figure}

\section{Model}

Our model consists of two components: a \fm{grounding module} and a
\fm{semantic module}. The grounding model processes an input image
using our novel parallel attention mechanism to
produce a set of $n$ \fm{grounding variables}
$\{ v_i\}_{1\leq i \leq n}$ each of which is meant to represent an
object, group of objects, or other groundable part of a scene. In
order to allow for differing numbers of objects across images, the
grounding module also returns a vector $h \in [0,1]^n$ of
\fm{objecthood parameters} which allows the model to turn off or on
particular grounding variables---allowing our model to learn the
number of objects in a scene from data.

The semantic module represents individual word denotations as well as
how these are composed. The meaning of CLEVR questions in our model is
represented using a domain specific language (DSL) derived from that
of \citet{mao.j:2019} and related to the DSL reported in the original
CLEVR paper \citep[][]{johnson.j:2017}.

At a high level, our semantics, like that of \citet{mao.j:2019}, makes
use of two leading ideas. First, the denotations of individual
predicates used in CLEVR questions, such as \ex{red} or \ex{in front
  of}, are formalized in terms of \fm{concept embeddings} which are
compared to grounding variables to determine the degree to which an
object matches the predicate. Second, composition is accomplished
primarily through fuzzy variants of standard logical operators such as
\textsc{and} and \textsc{or}.  We describe the model in greater detail
below.

\subsection{The Grounding Module}

As input to the grounding module, CLEVR images are represented as
$c$-channel feature maps $I \in \mathbb{R}^{l \times w \times c}$
derived from the pre-trained ResNet proposed in
\citet{he2016deep}. %\footnote{$w=16$,$l=24$,$c=$ in the simulations below.}
In what follows, we will refer to each set of channels at a particular
feature map position, that is $I_{[l,w,:]}$, as a \fm{column}.

Each grounding variable is bound to information that is the
result of selecting information from $I$ weighted by a
variable-specific attention map. We first describe how these attention
maps are computed using our parallel attention mechanism.

We start by computing a
\fm{foreground map} $F \in [0,1]^{l \times w}$ of the image. This
foreground map is meant to distinguish potential objects from
background and is used to initialize the parallel attention
mechanism:

\begin{equation}
  F = \sigma(\textsc{cnn}_{F}(I)).
  \end{equation}

  We compute initial attention maps $A_i^0 \in [0,1]^{l \times w}$ for
  each grounding variable $v_i$ by (i) finding the $n$ local maxima of
  $F$ and (ii) pre-initializing a separate attention map $A^{\mathrm{pre}}_i$
  with weights given by a Gaussian centered on the position of the
  $i$-th local maximum in $F$ and (iii) applying a convolutional
  neural network to derive the initial attention
  $A_i^0=\sigma(\textsc{cnn}_{0}([I;\bar{F};\bar{A}^{\mathrm{pre}}_i]))$. Here
  and below the notation $\bar{X}$ for a matrix $X$ means:
  $\bar{X}=\log(X)$.

  Our parallel attention mechanism iteratively computes a \fm{scope}
  $R_i \in [0,1]^{l \times w}$ and an attention map
  $A_i^m \in [0,1]^{l \times w}$ for each grounding variable $v_i$
  using the following update equations for step $m$:

\begin{align}
  R^m_i &= \prod_{j \neq i} \left [ 1-A^{m-1}_i \right],\\
A^{m}_i &= \sigma(\textsc{cnn}_{m}([I;\bar{F};\bar{R}^{m-1}_i;\bar{A}^{m-1}_i])).
\end{align}

\noindent Here each scope represents the portion of the image
unattended to by the other attentions. All attentions are updated in
parallel on each step $m$. We update $s$ times and take $A^{s}_i$ to
be the final attention map for each grounding variable
$v_i \in \mathbb{R}^{c}$.

The grounding variable $v_i$  is bound to the 
foreground- and attention-weighted average of channels over the
feature map:

\begin{equation}
  v_i = \sum_{j,k} (F \odot A^{s}_i)_{[j,k]} I_{[j,k,:]},
\end{equation}
where $\odot$ denotes element-wise multiplication. Note that we
include the foreground map $F$ in this computation in order to allow
training signal to flow to this object, since our \textsc{max}-based
initialization procedure is non-differentiable.

For use with relational predicates, we also derive a pair embedding
$v_{ij}$ for each pair of objects $i$ and $j$ by pushing the
concatenated object embeddings through a dimensionality-reducing
linear map $P$.

\begin{equation}v_{ij} = P[v_i;v_j].\end{equation}

Finally, each grounding variable is associated with an objecthood
parameter $h_i \in [0,1]$ meant to represent the model's confidence
that the information in grounding variable $v_i$ represents a valid
binding:

\begin{equation}h_i =
  \sigma(\textsc{mlp}_h(\textsc{cnn}_{h}([I;\bar{F};\bar{A}^{s}_i;\bar{R}^{s}_i]))).
  \label{eq:objecthood}
\end{equation}

\subsection{Semantic Module}
Semantics in our model is handled by a DSL based on that of
\citet{mao.j:2019}. Here we give an overview of how denotations are
represented and meanings are computed in our system.  For further
details, please see the Appendix.  We organize our
discussion into three parts. First, we discuss how \ex{extensions} of
particular concepts such as \conc{red}, \conc{square}, or
\conc{behind} are represented as \fm{fuzzy sets} and computed.
Second, we discuss how such extensions are combined
compositionally. Third and finally, we discuss how CLEVR questions are
implemented as special-case top-level semantic operators.

\subsubsection{Computing Extensions}
\label{sec:extensions}
In CLEVR, objects have a number of \fm{attributes} such as \att{color}
and \att{shape} whose \fm{values} are \fm{concepts} such as \conc{red}
and \conc{cube}.\footnote{We also include a null concept for each
  attribute which can be associated with non-objects.} We first
discuss how we compute \fm{extensions}---that is, representations of
the sets of objects that are consistent with some one- or two-place
predicate (e.g., the set of red things).  Extensions are represented
in our system as \fm{fuzzy sets}. A fuzzy set is a vector
$x \in [0,1]^n$ where each component $x_i$ indicates fuzzy set
membership for object $i$.

To compute the extension of a single-place predicate such as the
concept \conc{red} we construct a matrix 
$M \in [0,1]^{|\att{a}| \times n}$ in two steps. First, we compute a
matrix $M^\prime \in \mathbb{R}^{|\att{a}| \times n}$:

\begin{equation}
  M^\prime_{[\conc{m},i]} = y_{\conc{m}} \cdot A_{\att{a}} v_i,
 \end{equation}

 \noindent where $|\att{a}|$ is the number of values for the relevant
 attribute (e.g., \att{color}), $A_{\att{a}}$ is a linear
 transformation which moves the object embedding $v_i$ into attribute
 space and $y_{\conc{m}}$ is a concept embedding, for example
 $y_{\conc{red}}$. Thus, one-place predicates are evaluated by
 comparison (via dot product) of a grounding with a concept embedding.
 We further assume that concepts are mutually exclusive for
 a given object and, thus, perform a \textsc{softmax} operation on
 each column of $M^\prime$ to derive $M$. A row $\conc{m}$ in $M$
 represents a fuzzy set over objects for which the concept $\conc{m}$
 is true.

To compute the extension of a two-place spatial relation such as the
concept \conc{behind}, we construct a matrix
$O^{\conc{m}} \in [0,1]^{n \times n}$ of probabilities that each pair of objects
stands in the relation.

\begin{equation}
  O^\conc{m}_{[i,j]} = \sigma(zy_{\conc{m}} \cdot v_{ij})
\end{equation}

\noindent Here $y_{\conc{m}}$ is a concept embedding for the relation
(e.g., $y_{\conc{behind}}$) and $z$ is a scaling parameter. A row $i$
in $O^{\conc{m}}$ represents a fuzzy set over objects that stand in the target
relation $\conc{m}$ with the object $i$.

To compute the extension of a two-place attribute-identity 
predicate such as \ex{same \att{color} as \textsf{X}} we first
construct a concept-object matrix $M$ for the target attribute \att{a}
as above. We next compute a matrix $E^a \in [0,1]^{n \times n}$ of
probabilities that each pair of objects share a concept value. The probability that objects $i$ and $j$ share a
value for an attribute is given by exponentiating the negative
KL-divergence between their distributions over values of the
attribute; formally,

\begin{equation}
  E^a_{[i,j]} = e^{- \mathrm{KL(M_{[:,i]} || M_{[:,j]})}}.
\end{equation}

\noindent A row $i$ in $E^a$ represents a fuzzy set of objects that
have the same value for attribute \att{a} as object $i$.

\subsubsection{Combining Extensions}
\label{sec:combining}
In our system, there are two main mechanisms for combining
extensions. First, extensions can be combined by logical \textsc{and}
and \textsc{or}, for example, \ex{blue cube} can be represented as
\textsc{blue}(x) $\wedge$ \textsc{cube}(x) and \ex{blue or red thing}
can be represented as \textsc{blue}(x) $\vee$ \textsc{red}(x). To take
the intersection of two extensions, we use fuzzy \textsc{and}
implemented as a component-wise \textsc{min} operation on a pair of
fuzzy sets \citep[see,][]{ross.t:2017}. For disjunction, we
use fuzzy \textsc{or} implemented as a \textsc{max}.

In our implementation, all \textsc{and} operations are also passed the
objecthood parameters vector $h$ in order to ensure that grounding
variables which are not bound to objects do not participate in the
semantics. By threading these objecthood parameters through the
semantics in this way, we provide the model with a training signal for
the number of objects in a scene.

Second, CLEVR questions often contain definite noun phrases (e.g.,
\ex{the blue cube}). We normalize fuzzy sets to capture the uniqueness
presupposition associated with definites. For example, \ex{the blue
  cube} is represented by the probability vector which results from
normalizing the fuzzy set associated with \ex{blue cube}.  
%Let $p$ be
%the probability vector representing some definite noun phrase
%\texttt{NP}.  To compute the fuzzy set $x$ of objects that stands in a
%two-place relation with \texttt{NP} (e.g., \ex{\textsf{x} to the left
%  of [the blue cube]$_{\texttt{NP}}$}), we take $x = O^{\conc{m}}p$ or $x %= E^a p$
%where $O^{\conc{m}}$ and $E^a$ are the matrices defined in
%\textsection\ref{sec:extensions}.

\subsubsection{Top-Level Operators and Objectives}
CLEVR questions come in several varieties, each of which is
implemented by a unique top-level operator in our DSL. The top level
operators in our system are (i) \op{query} which takes a fuzzy set of
objects and an attribute and returns a probability distribution over
concepts associated with the attribute; (ii) 
\op{query-attribute-equality} which takes two definite noun phrases
and returns the expected probability that the two definite noun
phrases have the same attribute; (iii) \op{count} which takes a
fuzzy set and returns the sum (i.e., the expected cardinality of the
set); (iv) \op{exists} which takes a fuzzy set and returns the
max of the set which is the probability that the
existential is true; and (v) the three count-comparison
operators which each take two fuzzy sets as input and return  the
probability that the cardinality of the first set is greater than,
less than, or equal to the first. All operators use a cross-entropy
loss except for \op{count} which uses a squared error loss.

\section{Model Ablations and Variants}
\label{sec:ablations}

In order to better understand the behavior of our model, we performed several experiments on  variant models which  modify or ablate features of the architecture. We describe these in this section.

\subsection{Ablating Foreground Initialization}

Our model uses a foreground map in order to initialize the object attention maps; it initializes the object attentions at the $n$ local maxima of the foreground. We perform an ablation on this initialization strategy, instead randomly initializing the object attention maps by placing Gaussians centered at randomly chosen locations.

\subsection{Ablating Attention Scopes}

Our parallel attention mechanism makes use of scopes representing the portion of the image not attended to by the other grounding variables.  We ablate attention scopes by removing them from all computations. Thus each grounding variable is computed independently; variable $v_i$ does not receive any information about the location of the other objects.

\begin{align*}
A^{m}_i &= \sigma(\textsc{cnn}_{m}([V;\bar{F};\bar{A}^{m - 1}_i])) \\
h_i &=  \sigma(\textsc{mlp}_h(\textsc{cnn}_{h}([I;\bar{F};\bar{A}^{s}_i)))
\end{align*}

\subsection{Sequential Model Variant}

Our model assumes that all attentions are updated wholly in parallel and symmetrically. We relax this assumption and break the symmetry in attentional updates in two ways.

We first consider a variant of the model in which the attention scope for the $i$th object $v_i$ is computed based only on objects with a lower index. This is analogous to the difference between a transformer encoder (our original model) and decoder (sequential variant). More precisely, the scope $R_i^m$ for the $i$'th object is computed by:

\begin{equation*}
 R^{m}_{0} = 1, ~~~ R^{m}_i = \prod_{j=0}^{i-1} \left [ 1-A^{m}_j \right].  \end{equation*}

\subsection{Recurrent Model Variant}

We consider a second sequential model variant, in which attentions are computed recurrently. In this variant, the scope $R_i^m$ is computed from preceding attentions at the same step, as in the sequential model above. However, the attention map $A_i^{m}$ for object $i$ at layer $m$ is also computed from the scopes at the same step, rather than the preceding step. Formally, $R^{m}_i$ is computed as in the sequential model above and attention maps are computed as:

\begin{align}
%    R^{m}_{0} &= 1\\
%    R^{m}_i &= \prod_{j=0}^{i-1} \left [ 1-A^{m}_j \right] \\
    A^{m}_i &= \sigma(\textsc{cnn}_{m}([V;\bar{F};\bar{R}^{m}_i;\bar{A}^{m-1}_i])).
\end{align}

\section{Simulations}
Models were optimized using the AdaBelief optimizer \citep{zhuang2020adabelief}, with a learning rate of $4\cdot10^{-5}$ and batch size of 24. Each model was trained on a single RTX 2080 Ti. Following \citet[][]{mao.j:2019}, we used a curriculum learning strategy which incrementally introduced scenes and questions in the order of their complexity; scene complexity was measured by number of objects and question complexity by length of semantic parse and whether they included two-place spatial relations. All simulations assume access to ground truth semantic parses.

\section{Evaluation on CLEVR}

We first evaluate the model on the CLEVR dataset, a popular synthetic benchmark for question answering based on visual scenes of 3D-rendered objects \citep{johnson.j:2017}. The CLEVR training set consists of 70,000 images and 700,000 question/answer pairs, together with a validation set of 15,000 images and 150,000 question/answer pairs.

We evaluated the performance of our model on CLEVR in two ways. First, we evaluate accuracy on the question-answering task. Second, we evaluate whether the model is able to recover the correct attributes and spatial relations for objects in the scene---that is, whether the model can recover the  correct \fm{scene graph} of the image.

\begin{figure}
    \includegraphics[width=0.35\paperwidth]{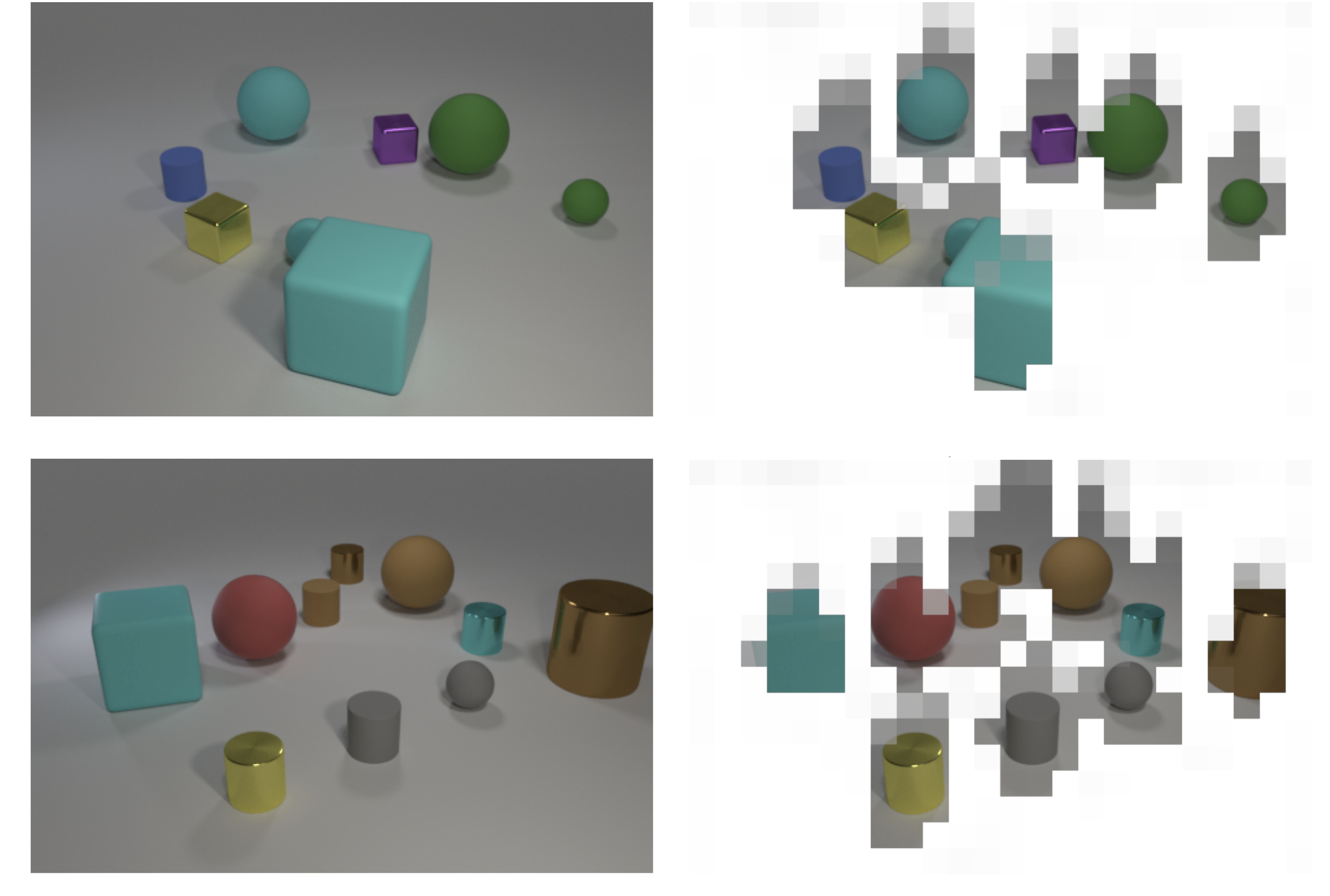}
    \caption{Attentions for the Parallel Attention model. Right panel shows sum total of model attentions in the scene. See Appendix for additional visualizations.}
    \label{fig:object-detection}
\end{figure}

\begin{table}[t]
\small
\begin{tabular}{lc}
\hline
\toprule
\multicolumn{2}{c}{\textbf{CLEVR Question Answering Performance}} \\ \midrule
\multicolumn{1}{c}{\textbf{Model}}       & \textbf{Accuracy}      \\ \midrule
Parallel Attention (PA)                 & 98.9 \\                  
PA-Sequential                              & 99.4                   \\ 
PA-Recurrent                               & 81.9                   \\ 
Ablate Initialization                      & 67.9                   \\ 
Ablate Scope                               & 97.5                   \\ 
\midrule
% you guys know this one
% notably they sort of also learn the program generator
% but of course they start from pre-detected objects
Neuro-Symbolic Concept Learner \\ \citep{mao.j:2019} & 99.2 \\
% this is the best attention-based neural module network
% it is very carefully designed to perform very well on CLEVR
% it uses ground-truth programs, but it does not rely on pre-detected objects
Transparency by Design \\ \citep{mascharka_transparency_2018} & 99.1 \\ 
% this is a very sophisticated attention-based recurrent model, trains completely from scratch
Compositional Attention Networks \\ \citep{hudson_compositional_2018} & 98.9 \\ 
% this is the best you can get with the vanilla deep stack of layers
Feature-Wise Linear Modulation \\ \citep{perez_film:_2017} & 97.7 \\ 
\bottomrule
\end{tabular}
\caption{Question accuracy on the CLEVR validation set.}
\label{tab:clevr-accuracy}
\end{table}

% \begin{table}[]
% \begin{tabular}{|l|c|}
% \hline
% \multicolumn{2}{|c|}{\textbf{CLEVR Question Answering Performance}} \\ \hline
% \multicolumn{1}{|c|}{\textbf{Model}}       & \textbf{Accuracy}      \\ \hline
% Parallel Attention (PA)                 & 98.9                   \\ \hline
% PA-Sequential                              & 99.4                   \\ \hline
% PA-Recurrent                               & 81.9                   \\ \hline
% Ablate Initialization                      & 67.9                   \\ \hline
% Ablate Scope                               & 80.5                   \\ \hline
% \end{tabular}
% \caption{Question accuracy on the CLEVR validation set.}
% \label{tab:clevr-accuracy}
% \end{table}

Table~\ref{tab:clevr-accuracy} shows model accuracy on the CLEVR validation set for our model, the ablated/variants models discussed in \textsection\ref{sec:ablations}, and a number of state of the art models from the literature. As can be seen in the table, our base model (Parallel Attention) achieves state-of-the-art performance on CLEVR question answering. Interestingly, the best performance is exhibited by the sequential variant of our model, which we discuss below.

\begin{table*}[]
\small
\centering
\begin{tabular}{lcccc}
\toprule
\multicolumn{5}{c}{\textbf{CLEVR Scene Graph Performance}}                                                                                                                                      \\ \midrule
\multicolumn{1}{c}{\textbf{Model}} & \textbf{Att-Precision} & \multicolumn{1}{l}{\textbf{Att-Recall}} & \multicolumn{1}{l}{\textbf{Rel-Precision}} & \multicolumn{1}{l}{\textbf{Rel-Recall}} \\ \midrule
Competitive Attention (CA)           & 99.8                   & 99.3                                     & 99.4                                        & 98.9                                     \\ 
CA-Sequential                        & 99.8                   & 99.6                                     & 99.7                                        & 99.4                                     \\ 
CA-Recurrent                         & 98.9                   & 98.4                                     & 58.8                                        & 58.6                                     \\ 
Ablate Initialization                & 93.9                   & 74.9                                     & 62.3                                        & 49.5                                     \\ 
Ablate Scope                         & 99.2                   & 98.3                                     & 98.9                                       & 97.9                                     \\ 
\bottomrule
\end{tabular}
\caption{Model performance at recovery of the ground truth object attributes (Att) and relations (Rel) in validation set scenes.}
\label{tab:clevr-precision-recall}
\end{table*}

Table~\ref{tab:clevr-precision-recall} shows the results of the scene-graph evaluations on the CLEVR dataset. We report the model's ability correctly recover the values of particular attributes of individual objects (i.e., \conc{red} or \conc{cube}). Recall measures the percentage of correctly identified attribute values over the total number of gold standard object attribute values, and precision measures the percentage of correctly identified attribute values over the total number of predicted object attribute values. Similarly, we report precision and recall for spatial relations (i.e., \conc{behind} or \conc{left}).

As can be seen from Tables~\ref{tab:clevr-accuracy} and \ref{tab:clevr-precision-recall}, eliminating the foreground initialization leads to significant degradations in performance in both question answering and scene graph recovery; eliminating attention scopes leads to a smaller degradation. It appears to be important to allow each grounding variable attention to start from a specific local maximum in the foreground map and to adjust based on what the other grounding variables are looking at. Interestingly, this degradation in performance on scene graphs is much higher in terms of recall suggesting that without foreground initialization, the model is finding fewer of the correct objects. 
%By contrast, precision remains relatively high, suggesting that the model is able to correctly identify attributes and relations for objects it does find.

Our base model assumes that all attentions are updated completely in parallel and symmetrically; symmetry between grounding variables is only broken by the foreground initialization. The sequential and recurrent variants of the model relax this assumption and introduce an ordering into the parallel attention mechanism. Interestingly, the sequential model seems to improve the overall performance of the model, and gives what we believe is the current state of the art for CLEVR question answering. We leave exploring this architecture variant more thoroughly to future work. By contrast, the recurrent architecture leads to a substantial decrease in performance. We note however that this architecture exhibited unstable training performance, suggesting the recurrent connections led to gradient explosions.

\section{Linguistically Influenced Grounding}

\begin{table}[]
\small
\begin{tabular}{lcl}
\toprule
\multicolumn{3}{c}{\textbf{Non-Canonical Question Accuracy}}                             \\ 
\midrule 
\multicolumn{1}{c}{\textbf{Task}} & \textbf{All questions} & \textbf{Target questions} \\ 
\midrule
Ignore Red                          & 97.8                   & 90.9                        \\
Group Cubes                         & 98.8                   & 90.1                        \\ 
\bottomrule
\end{tabular}
\caption{Validation set accuracy for the non-canonical grounding tasks. Target questions are those which have a different answer in the modified validation set compared to the original CLEVR dataset.}
\label{tab:non-canonical-results}
\end{table}

As we discussed in the introduction, one of the motivations behind this work is to explore how it is possible to learn the tight contingencies between the way that language is used and the kinds of groundings that are available in a particular situation. To explore this question, we perform several experiments to illustrate the top-down influence of linguistic usage on the kinds of grounding that our model learns. In each of these experiments, we train the system to acquire some non-canonical grounding that goes beyond the simple object groundings of the original CLEVR dataset. These experiments make use of the same set of scenes and questions as the standard CLEVR dataset---only the answers to questions are modified.  Thus, the linguistic input provides the only training signal for the non-canonical groundings.

\begin{figure}
    \includegraphics[width=0.35\paperwidth]{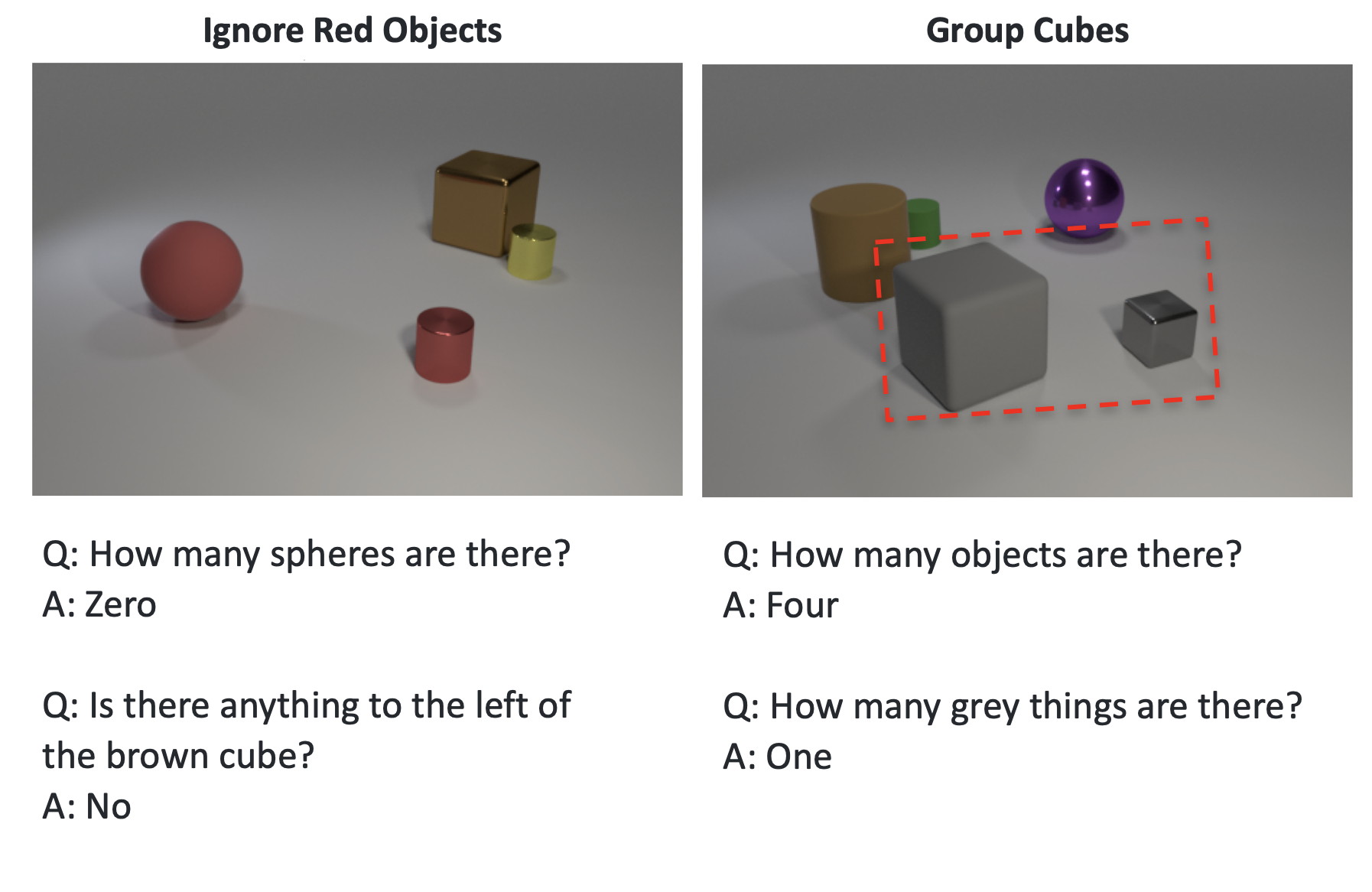}
    \caption{This figure illustrates the training setup for our linguistically influenced grounding experiments. The red bounding box is for illustration purposes only.}
    \label{fig:deobjectify-training}
\end{figure}

\subsection{Learning to Ignore Red Objects}\label{sec:ignore-red}

What counts as background in a scene depends on situational and linguistic context. In the introduction, we gave the example of the objects depicted in a painted mural which might be discussed in a lecture on art history, or ignored when discussing how to rearrange furniture in a room. As a simple analogue in CLEVR, we examine whether the model can learn to treat all red objects as background.

The model is trained and evaluated on a modified version of CLEVR. All scene images remain unchanged. However, the ground-truth answers to questions are modified so that any red objects in the scene are ignored. This is illustrated in the left panel of Figure \ref{fig:deobjectify-training}; the scene and questions are taken from the original dataset, but the answers have been updated. Questions containing the word \ex{red} are removed.

\begin{figure}
    \includegraphics[width=0.35\paperwidth]{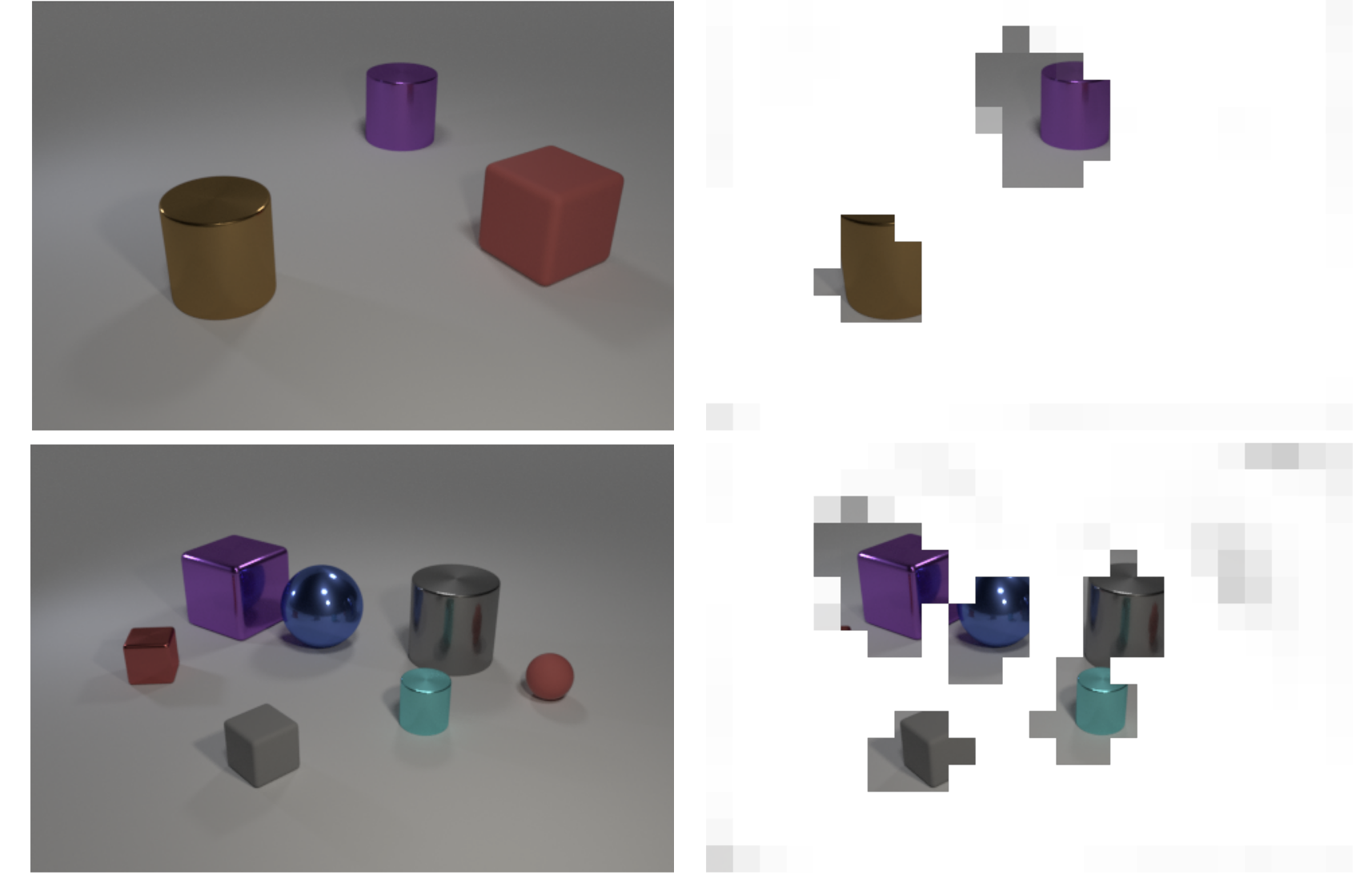}
    \caption{Attentions for the model trained to ignore red objects. Right panel show the sum total of model attentions in the scene.}
    \label{fig:ignore-red}
\end{figure}

Table~\ref{tab:non-canonical-results} shows validation performance for the task. Results are shown for all questions in the validation set, and for target questions whose answer had changed compared to original CLEVR.\footnote{Some answers did not change in the new dataset, for example, in scenes that did not contain any red objects.} As can be seen from the table, the model is able to learn that red objects are part of the background and should not be bound to objects. We emphasize that the model was able to learn this generalization just from changing the answer to relevant questions.

Figure~\ref{fig:ignore-red} shows the summed attentions over all grounding variables for several scenes for the trained model. As can be seen from the figure, the model has learned to ignore all red objects.

\subsection{Learning to Group Cubes}

\begin{figure}
    \includegraphics[width=0.35\paperwidth]{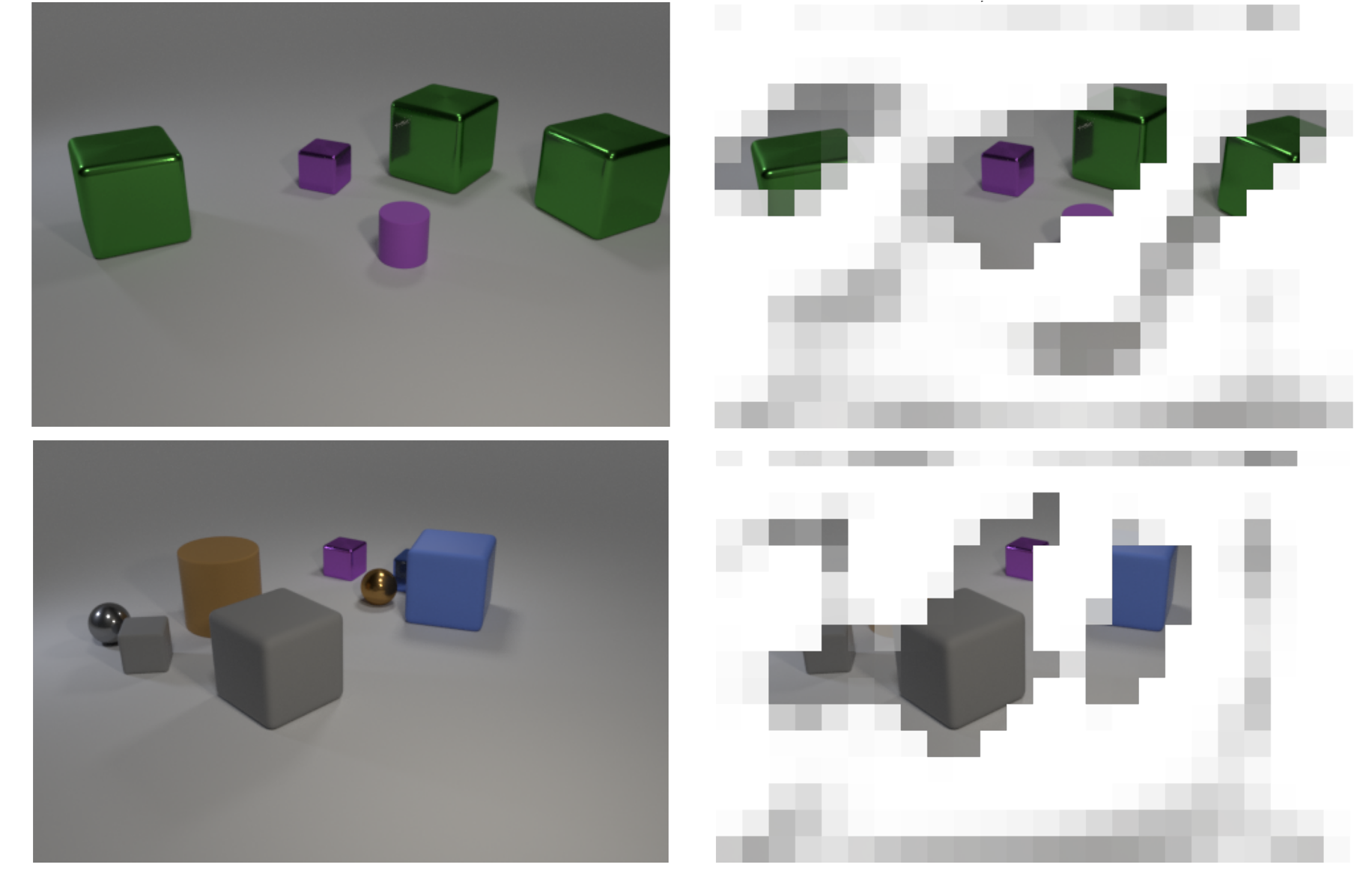}
    \caption{Example attentions for the model trained to group cubes. The right panel shows attention for the single \conc{cube}-valued grounding in the scene; this corresponds to a single grounding variable $v_i$ in the model. }
    \label{fig:group-cubes}
\end{figure}

In a second experiment, we evaluate whether the model can learn to treat a group of objects as a single, linguistically-indexable object. We develop a new task, in which the set of cubes in a scene is treated as a single object for the purposes of questions and answers. All scene images remain unchanged from CLEVR. We create training and validation sets of zero-hop questions (which includes query, count, and exists questions, and excludes two-place spatial relations). Answers to these questions are computed from a scene representation in which all cubes are treated as a single object.\footnote{The cube group object inherits any properties that all of its constituents have. For example, in Figure \ref{fig:group-cubes}, the cube group is grey, since each of its constituent cubes is grey.} This is illustrated in the right panel of Figure \ref{fig:deobjectify-training}.

Validation performance is shown in Table \ref{tab:non-canonical-results}. Once again the model is able to learn that all cubes should be treated as a single object just from linguistic evidence alone.

Figure~\ref{fig:group-cubes} visualizes the attention associated with a single grounding variable that the model identifies as have the  \conc{cube} value of the \att{shape} attribute. This corresponds to the attention of the single object slot which the model categorizes as a cube.

% \begin{table}[]
% \begin{tabular}{|l|c|l|}
% \hline
% \multicolumn{3}{|c|}{\textbf{Non-Canonical Question Accuracy}}                             \\ \hline
% \multicolumn{1}{|c|}{\textbf{Task}} & \textbf{All questions} & \textbf{Target questions} \\ \hline
% Ignore Red                          & 97.8                   & 90.9                        \\ \hline
% Group Cubes                         & 98.8                   & 90.1                        \\ \hline
% \end{tabular}
% \caption{Validation set accuracy for the non-canonical grounding tasks. Target questions are those which have a different answer in the modified validation set compared to the original CLEVR dataset.}
% \label{tab:non-canonical-results}
% \end{table}

\section{Conclusion}

In this paper, we have introduced a model which jointly learns the
denotation of words and how to ground them in images. The model
extends the architecture of \citep[][]{mao.j:2019} with a novel
parallel attention mechanism which allows it to combine the advantages
of truth-conditional semantics and attention-based grounding in a
differentiable system.  We achieve state of the art performance on
question answering on the CLEVR dataset. We also show that model is
able to learn to ground the meaning of linguistic expressions in novel
and non-canonical ways based purely on changes to the linguistic
input. We hope that the model can serve as an early step in studying the
intricate problem of learning the contingencies between linguistic
meaning and grounding in a way that is able to take advantage of tools
from both machine learning and formal semantics.

\bibliography{neural-drs,grounding}
\bibliographystyle{acl_natbib}

\newpage

\appendix

The first appendix provides a description of the semantic operators used in the paper. The second appendix provides additional visualizations for the experiments.

\section{Semantic Operators}
\label{sec:appendix}

In this appendix we give a detailed breakdown of the implementation of
our semantic operators. Note that we use the same set of operators as
in Mao et al. (2019), while the detailed implementation of
several of the operators has changed.

Note that the definiteness transformation described in
\textsection{sec:combining} is performed within several of the
operators when these expect definite inputs. Similarly, several of the
operators perform \textsc{and} as a final step before output once an
appropriate extension has been computed. 

%filter changed, since we now assume attribute values are mutually exclusive
%relate attribute equality and query attribute equality have changed

\subsection{Filter}

\begin{center}\ex{blue \textsf{X}}\end{center}

The \op{filter} operator filters a set of objects based on some
concept such as \conc{red} or \conc{cylinder}. It takes as inputs (i)
a specification of the concept (and corresponding attribute such as
\att{color} or \att{shape}); (ii) the vector of objecthood parameters
$h_i$ (iii) a vector $x \in [0,1]^n$ which representing the extension
resulting from earlier semantic computation, and (iv) the set of
object grounding vectors $v_i$.

The \op{filter} operator first computes a matrix of representing the
degree to which each object fits each concept
$M^\prime \in \mathbb{R}^{|\mathcal{A}| \times n}$.

$$M^\prime_{[\conc{m},i]} = y_{\conc{m}} \cdot A_{\att{a}} v_i $$

Here $A_{\att{a}}$ is a linear transformation which moves the object
embedding $v_i$ into attribute space (e.g., \att{color}) and
$y_{\conc{m}}$ is a concept embedding for example $y_{\conc{red}}$.

We assume that concepts are mutually exclusive for a given object. To
capture this we perform a \textsc{softmax} operation on each column of
$M^\prime$ giving the resulting matrix $M$. 

Finally, we retrieve the row of $M$ corresponding to the target
concept and take an elementwise \textsc{min} operation between this
row, the input fuzzy set $x$ and the the objecthood parameters
$h$. This final \textsc{min} step represents the composition of the
results of this filter operation with earlier semantic operations.

\subsection{Relate}

\begin{center}\ex{\textsf{X} to the left of the blue cube}\end{center}

The \op{relate} operator computes the set of objects that stand in
some spatial relation with some target object. It takes as input (i)
the spatial relation (ii) a fuzzy set $\{x_i\}$ representing the
target object (iii) the set of object pair-embeddings
$\{ v_{ij}\}$ and (iv) the set of $\{h_i\}$ objecthood
parameters.

In the CLEVR database, all target objects in relations are definites
\ex{... to the left of \textbf{the cube}.} To capture the uniqueness
presupposition of definite noun phrases, we first renormalize the
target object vector so that it represents a probability distribution
over target objects $p = \textsc{renormalize}(x)$.

We next compute a matrix $O \in [0,1]^{n \times n}$ of probabilities
that each pair of objects stands in the relation.

$$O_{[i,j]} = \sigma((y_{\conc{m}} \cdot v_{ij}) \times z) $$

\noindent where $y_{\conc{m}}$ is a concept embedding for the relation
(e.g., $y_{\conc{behind}}$) and $z$ is a scaling parameter.

We then compute the joint probability that each object stands in the
relation with an another object  and that other object is the target
object as $t=Op$.

Finally, we take an elementwise \textsc{min} operation between each
value $t_i$ and the objecthood
parameters $h_i$.

\subsection{Relate Attribute Equality}

\begin{center}\ex{\textsf{X} that has the same color as the cube}\end{center}

The \op{relate-att-eq} operator computes the set of objects that
have the same concept value for some attribute as some target
object. It takes as input (i) the attribute (ii) a fuzzy set $\{x_i\}$
representing the target object (iii) the set of object groundings
$\{ v_{i}\}$ and (iv) the set of $\{h_i\}$ objecthood parameters.

As for \op{relate}, all target objects for attribute equality are
definites. Thus, we first renormalize the target object vector so that
it represents a probability distribution over target objects
$p = \textsc{renormalize}(x)$.

The \op{relate-att-eq} operator first computes a matrix $M$
representing the degree to which each object fits each concept in
precisely the same way as in the \op{filter} operator. We next compute
a matrix $O \in [0,1]^{n \times n}$ of probabilities that each pair of
objects share a concept value.

$$O_{[i,j]} = e^{- \mathrm{KL(M_{[:,i]} || M_{[:,j]})}}$$

\noindent We assume here concept values are mutually exclusive for a
particular attribute (e.g., \conc{red} or \conc{blue} for \att{color})
by applying \textsc{softmax} over each column of $M$. We compute the
probability that objects $i$ and $j$ share a concept value by
exponentiating the negative KL-divergence between their concept
distributions.

We then compute the joint probability that each object stands in the
relation with an another object and that other object is the target
object as $t=Op$.

Finally, we take an elementwise \textsc{min} operation between each
value $t_i$ and the objecthood parameters $h_i$.

\subsection{Intersect}

\begin{center}\ex{red cubes and green cylinders.}\end{center}

The \op{intersect} operator takes in two fuzzy sets and computes the
component wise \textsc{min} operation between them.

\subsection{Disjunction}

\begin{center}\ex{either small cylinders or metal things}\end{center}

The \op{disjoins} operator takes in two fuzzy sets and computes the
component wise \textsc{max} operation between them.

\subsection{Query}

\begin{center}\ex{What's the color of the cube}\end{center}

The \op{query} operator computes the probability that a target object
takes on some value of an attribute. It takes as input (i) the
attribute (ii) a fuzzy set $\{x_i\}$ representing the target object
(iii) the set of object groundings $\{ v_{i}\}$ and (iv) the set of
$\{h_i\}$ objecthood parameters.

As above, all target objects for attribute equality are
definites. Thus, we first renormalize the target object vector so that
it represents a probability distribution over target objects
$p = \textsc{renormalize}(x)$.

The \op{query} operator then computes a matrix representing the degree
to which each object fits each concept
$M \in [0,1]^{|\mathcal{A}| \times n}$ in exactly the same way as
\op{filter}.

We then compute vector of joint probabilities that the each object is
the target object and has the particular concept value. We renormalize
this vector to derive the desired conditional probability distribution
over concept values for the attribute and object
$o= \textsc{renormalize}(p^\intercal M)$.

\subsection{Count}

\begin{center}\ex{How many cubes are there?}\end{center}

The \op{count} operator takes a fuzzy set and returns the sum of the
values of the set, that is, the expected cardinality of the set.

\subsubsection{Exists}

\begin{center}\ex{Is there a red cube?}\end{center}

The \op{exists} operator takes a fuzzy set and returns the max of the
set. 

\subsection{Count Greater/Less/Equal}
\begin{center}\ex{Are there more red objects than gray blocks?}\end{center}

The three count comparison operators takes two fuzzy sets as
input. They compute the sum of each set $s_1$ and $s_2$ and then
compute the fuzzy truth value of the result according to one of the
following equations.

\begin{itemize}
\item \textbf{greater-than} $\sigma(A_{>}(s_1-s_2) + b_{>})$
\item \textbf{less-than} $\sigma(A_{<} (s_1-s_2) + b_{<})$
\item \textbf{equal-to} $\sigma(-A_{=}|s_1-s_2|  + b_{=})$
\end{itemize}

\subsection{Query Attribute Equality}
\begin{center}\ex{Do the block and sphere have the same color?}\end{center}

The \op{query-attribute-equality} operator takes fuzzy sets as
input. It performs a definiteness normalization on both then passes
one to the \op{relate attribute equality} operator. It then returns
the dot product of the fuzzy set returned by this operator and the
other definite noun phrase. This represents the expected probability
that two objects have the same value for some attribute.

%During training, this operator uses a binary cross-entropy
%loss. During test decoding we round the result returned by the process
%above.

\section{Attention Visualizations}

\begin{figure*}[h]
\centering
    \includegraphics[width=0.45\paperwidth]{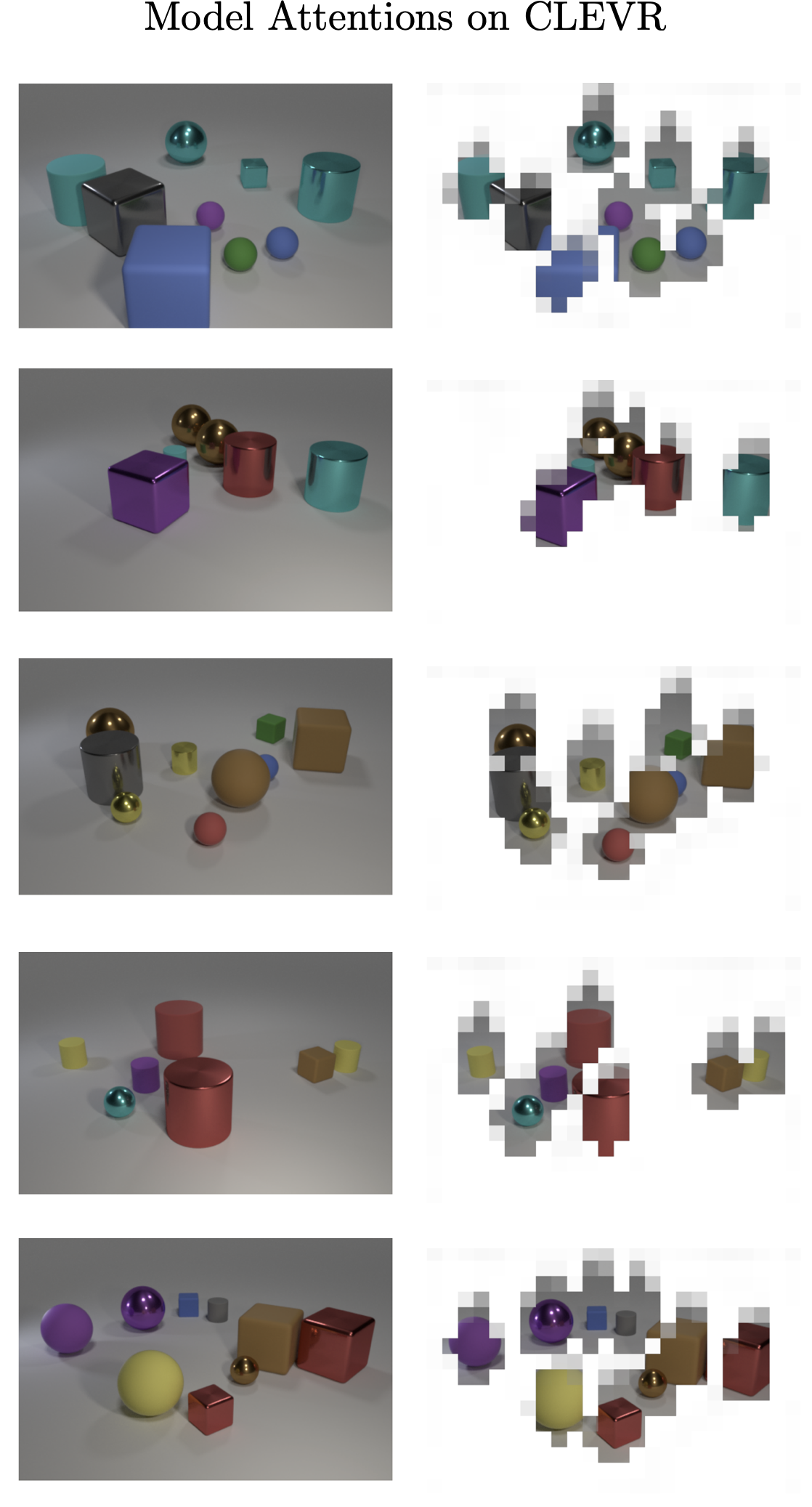}
    \caption{Attentions for the Parallel Attention model. Right panel shows sum total of model attentions in the scene. }
    \label{fig:object-detection-appendix}
\end{figure*}

\begin{figure*}[h]
\centering
    \includegraphics[width=0.45\paperwidth]{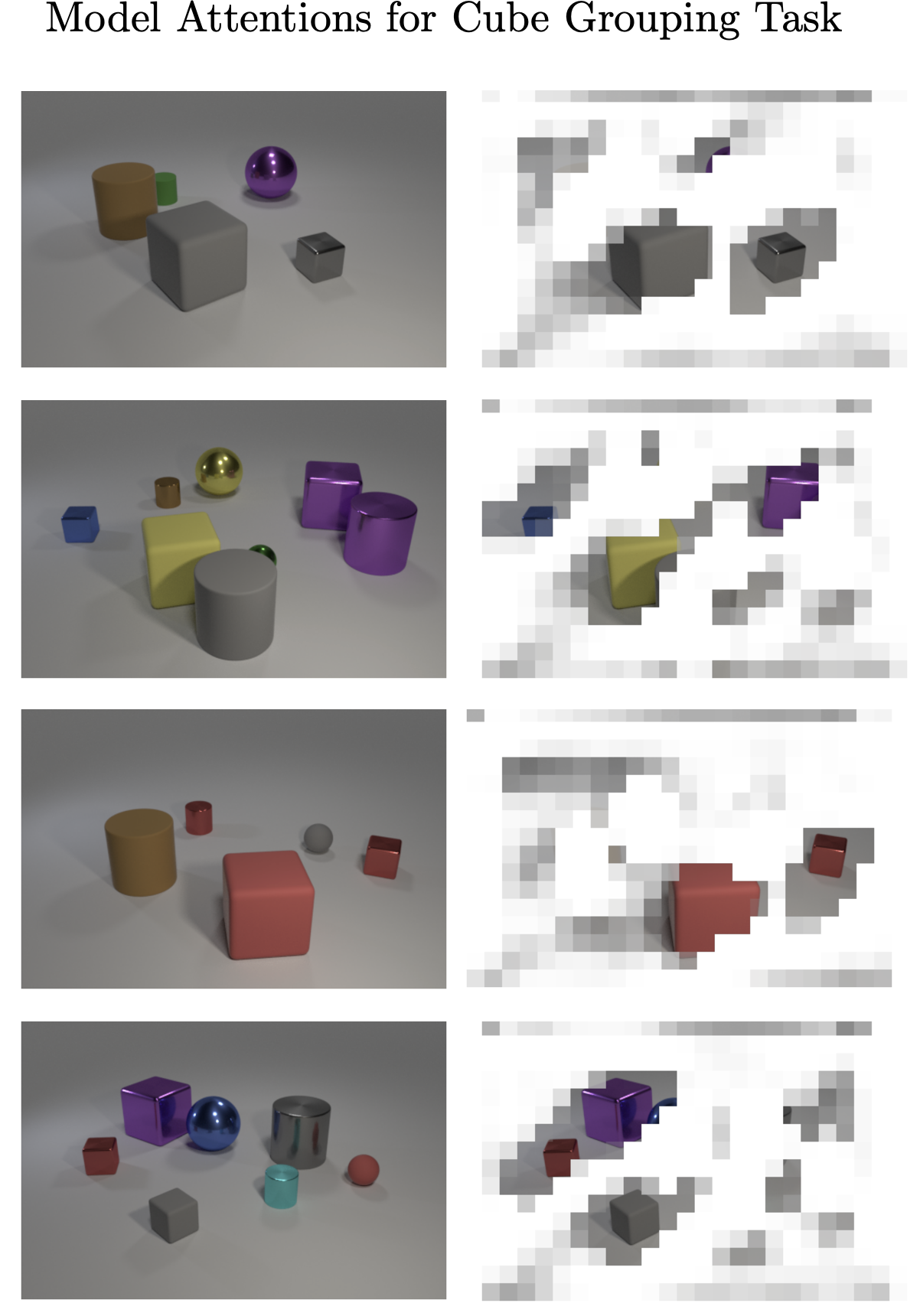}
    \caption{Attentions for the cube grouping task. Right panel shows the attention for the single object that the model categorizes as a cube. }
    \label{fig:group-cubes-appendix}
\end{figure*}

\begin{figure*}[h]
\centering
    \includegraphics[width=0.45\paperwidth]{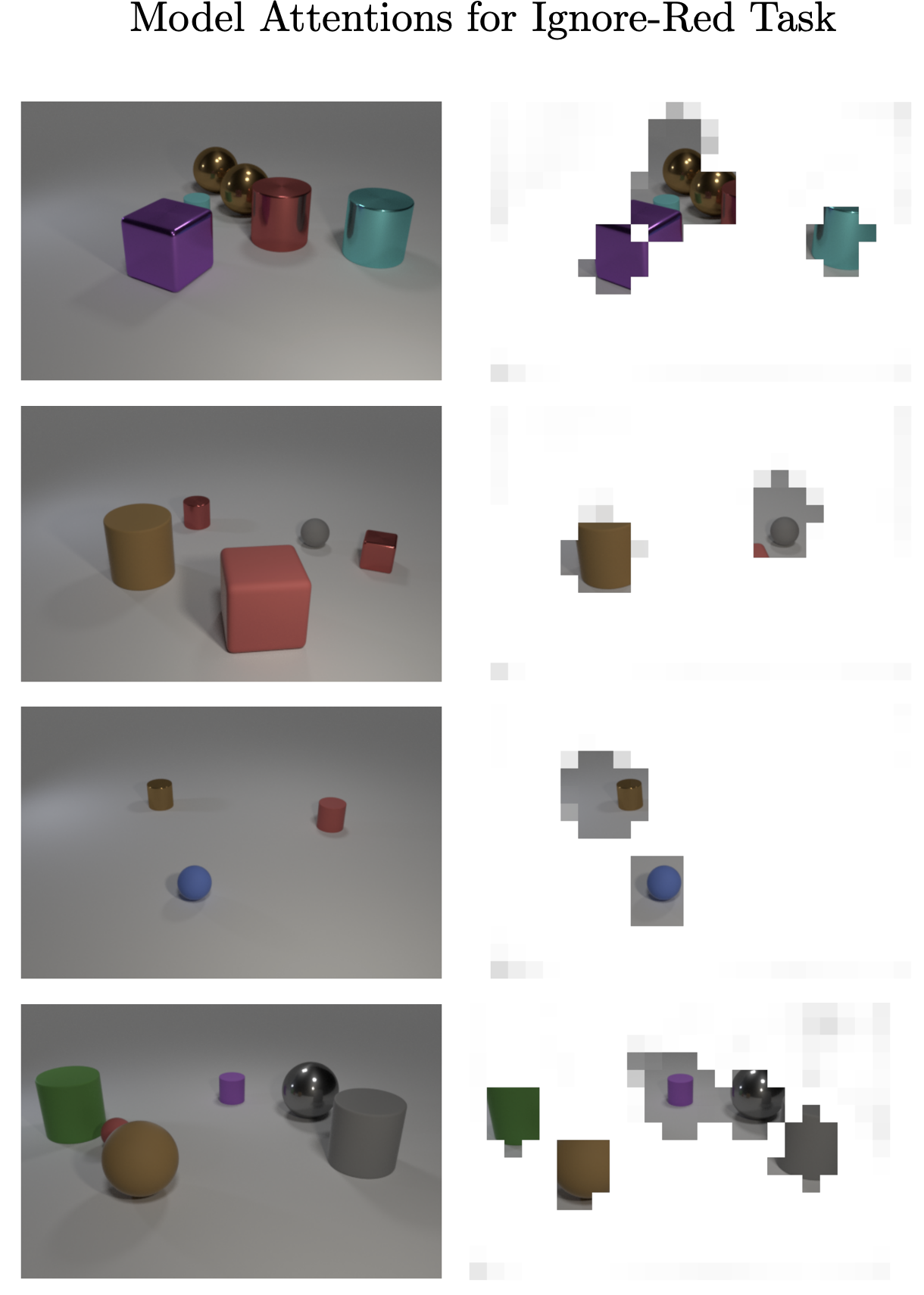}
    \caption{Attentions for the ignore-red task. Right panel shows the sum total of model attentions in the scene. }
    \label{fig:ignore-red-appendix}
\end{figure*}

\end{document}